\documentclass[lettersize,journal]{IEEEtran} % Comment this line out if you need a4paper

\IEEEoverridecommandlockouts              
\usepackage{xcolor}
\long\def\invis#1{}

\usepackage{amssymb}
\usepackage{mathtools}
\usepackage{graphicx}
\usepackage{float}
\usepackage{tikz}
\usepackage{listings}
\usepackage{hyperref}
\usepackage{graphicx}
\usepackage{cite}
\usepackage{dsfont}
\usepackage{mathtools,etoolbox}
\usepackage[ruled, linesnumbered]{algorithm2e}
\usepackage[none]{hyphenat}
\usepackage[utf8]{inputenc}
\usepackage[english]{babel}
\usepackage[T1]{fontenc}
\usepackage{mathtools, nccmath}

\usepackage{amsthm}
\usepackage{xfrac}
\usepackage{nicefrac}
\usepackage{booktabs}
\usepackage[switch, pagewise]{lineno}
\usepackage{mathrsfs}
\DeclareMathAlphabet{\mathpzc}{OT1}{pzc}{m}{it}
\usepackage{multirow}
\usepackage{multicol}
\usepackage{bbm}
\usepackage{soul}
\usepackage{xfrac}
\usepackage{balance}
\usepackage{titlesec}
\usepackage[final]{pdfpages}
\usepackage{amsmath}
\usepackage{xcolor}

\usepackage{cite}
\usepackage{amsmath,amssymb,amsfonts}
\usepackage{algorithmic}
\usepackage{graphicx}
\usepackage{mathtools}
\usepackage{textcomp}
\usepackage{xcolor}
\usepackage{url}
\usepackage[affil-it]{authblk}

\def\BibTeX{{\rm B\kern-.05em{\sc i\kern-.025em b}\kern-.08em
    T\kern-.1667em\lower.7ex\hbox{E}\kern-.125emX}}
\begin{document}

% \title{Simulation of Aerial and Satellite Images for Detection of Maritime Objects\\
\title{Whale Detection Enhancement through \\Synthetic Satellite Images\\
%\thanks{Corresponding author: Akshaj Gaur: agaur@umd.edu}
}

\author{Akshaj Gaur, Cheng Liu, Xiaomin Lin, Nare Karapetyan, Yiannis Aloimonos}
\affil{Maryland Robotics Center, University of Maryland}
\makeatletter
% \g@addto@macro\@maketitle{
% \setcounter{figure}{0}
% \vspace{-5mm}
% \begin{figure}[H]
%   \setlength{\linewidth}{\textwidth}
%   \setlength{\hsize}{\textwidth}
%     \centering
%     \includegraphics[width=\textwidth]{./figures/whale.png}
%     \vspace{-10mm}
%     \caption{First row is the real whale images taken from space and second row is the simulated whale images from Blender.}
%     \vspace{-11mm}
%     \label{fig:whale}
%     \end{figure}
% }
\maketitle
\thispagestyle{empty}
\pagestyle{empty}

%%%%%%%%%%%%%%%%%%%%%%%%%%%%%%%%%%%%%%%%%%%%%%%%%%%%%%%%%%%%%%%%%%%%%%%%%%%%%%%%
\begin{abstract}
With a number of marine populations in rapid
decline, collecting and analyzing data about marine populations has become increasingly important to develop effective conservation policies for a wide range of marine animals, including whales. \invis{Images of different domains (aerial, satellite, or underwater) can provide detailed information and wide coverage, enabling the assessment of environmental changes, studying biodiversity, and supporting conservation efforts for whales.} Modern computer vision algorithms allow us to detect whales in images in a wide range of domains, further speeding up and enhancing the monitoring process. However, these algorithms heavily rely on large training datasets, which are challenging and time-consuming to collect particularly in marine or aquatic environments. Recent advances in AI however have made it possible to synthetically create datasets for training machine learning algorithms, thus enabling new solutions that were not possible before. In this work, we present a solution - \textit{SeaDroneSim2} benchmark suite, which addresses this challenge by generating aerial, and satellite synthetic image datasets to improve the detection of whales and reduce the effort required for training data collection. We show that we can achieve a $15\%$ performance boost on whale detection compared to using the real data alone for training, by augmenting a $10\%$ real data. We open source \footnote{\url{https://github.com/prgumd/SeaDroneSim2}} both the code of the simulation platform \textit{SeaDroneSim2} and the dataset generated through it.\invis{By leveraging these synthetic images, computer vision systems are enhanced, contributing to improved monitoring and conservation of whale p ecosystems.}

\end{abstract}

\section{INTRODUCTION}
\label{sec:intro}
Satellite images play an increasingly crucial role in diverse tasks such as land-use classification \cite{chen2020imaging}, precision agriculture \cite{raj2020precision,tsouros2019review}, coastal management \cite{adade2021unmanned}, search and rescue missions \cite{albanese2021sardo}, and environmental monitoring \cite{cubaynes2019whales,gonccalves2020sealnet, karapetyan2021robot}. These images provide an accessible and comprehensive tool for marine monitoring, offering extensive coverage and detailed Earth surface insights. Remote sensing techniques using these images empower scientists and authorities to enhance marine ecosystem understanding, support conservation endeavors, and improve emergency response capabilities.
\begin{figure}
    \centering
    \includegraphics[width=1\columnwidth]{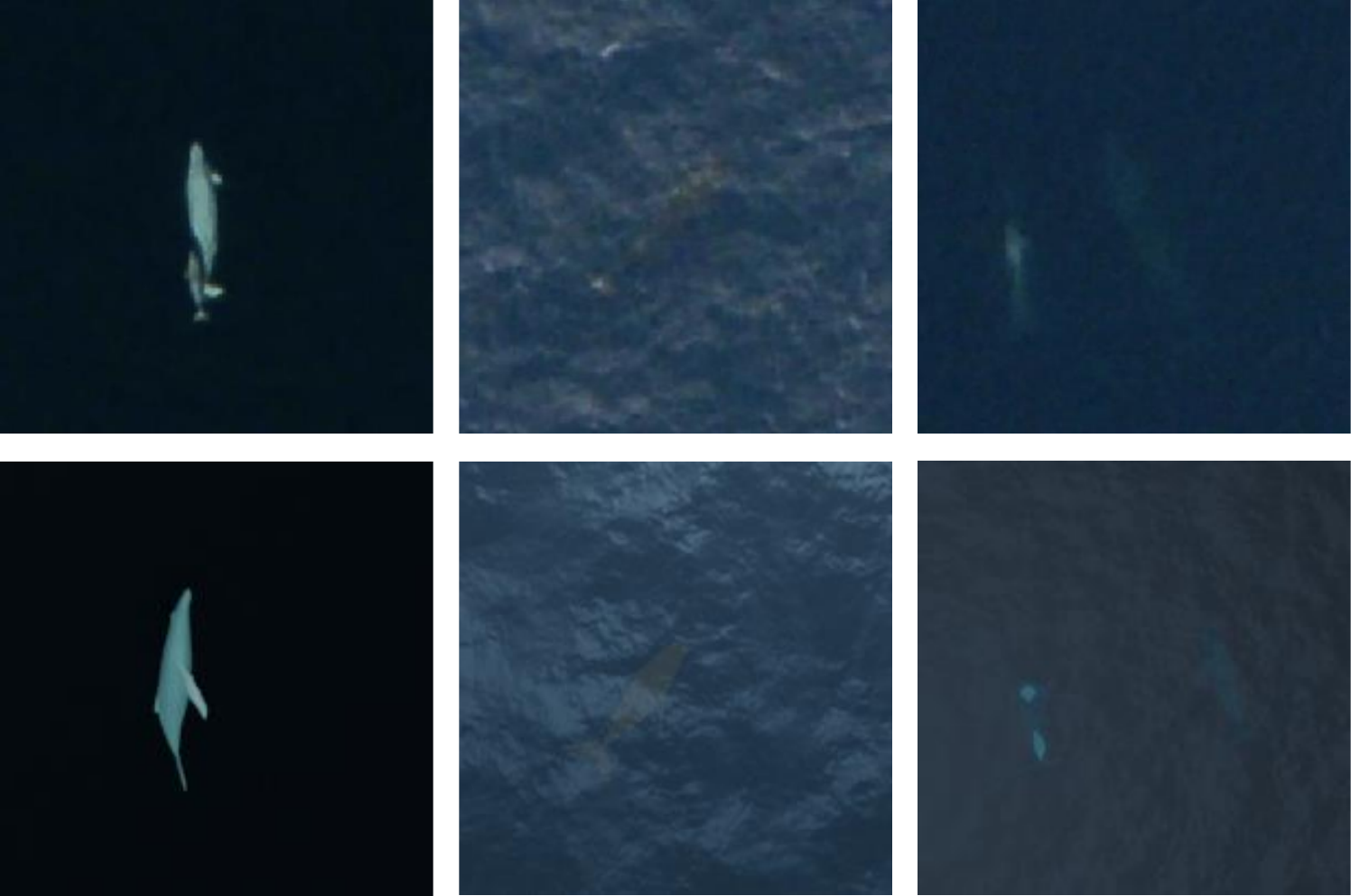}
    \caption{First row presents real whale images taken from space while the second row is the simulated whale images from using \textit{SeaDroneSim2}.}
    % \vspace{-5mm}
    \label{fig:illustrative_example}
\end{figure}

In maritime operations the significance of a dependable vision-based system cannot be overstated. In particular whale detection missions encounter various challenges such as lighting effects on mammals' visibility, image altitudes affecting appearances, underwater clarity reduction due to turbidity, varied watercolor grading, and diverse backgrounds. Additionally, object pose and texture variations are crucial for precise detection and tracking.

\begin{figure*}[ht!]
\includegraphics[width=\textwidth]{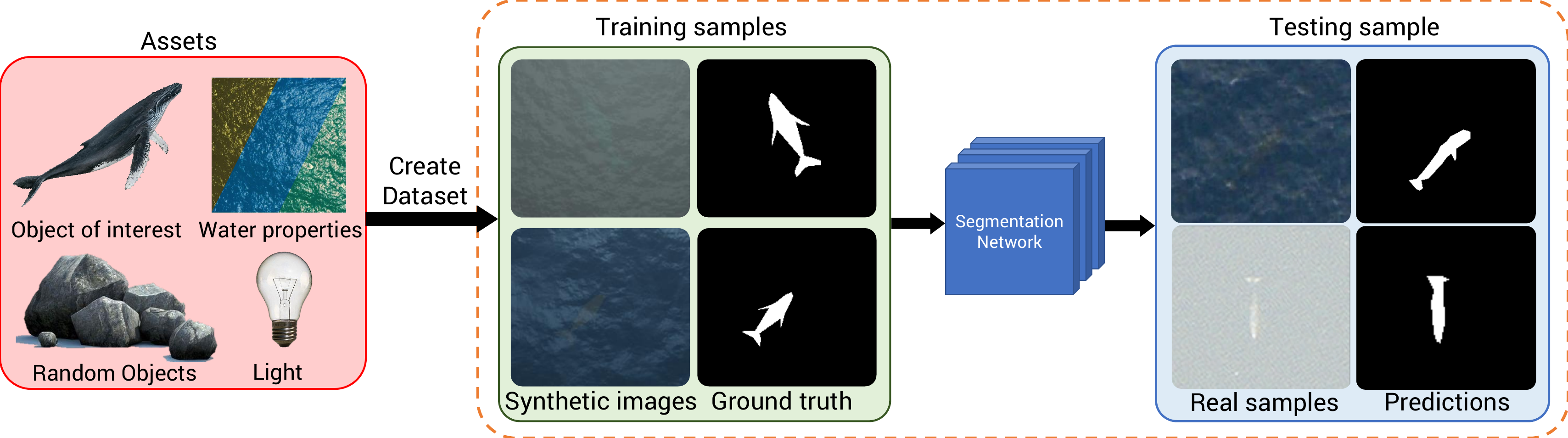}
\centering
\vspace{-3mm}
\caption{An overview of our approach. (a) Assets: Loads the assets such as water properties, objects, materials, etc. into  \textbf{\textit{SeaDroneSim2}} to generate Synthetic datasets. Note, the synthetic dataset would include its ground truth mask for the object of interest. (b) We modify the properties within the scene such as noise level, rotation of the object, altitude of the camera and etc. (c)The synthetic dataset generated is then fed into a Neural network to obtain the object detection result. We demonstrated the generation of aerial, satellite, and underwater images for two different objects of interest in our study. Note: Object Detection images are cropped and enlarged for better visualization.}
\label{fig:overview}
\vspace{-3mm}
\end{figure*}
To address these challenges, the integration of robust algorithms becomes imperative, particularly within the realm of deep neural networks. However, the availability of datasets for whale detection in maritime environments, despite some efforts from Cubaynes~\cite{cubaynes2022whales}, remains limited in terms of both size and diversity. Acquiring basic remote sensing images for smaller regions can cost over \$100,000 \cite{cost}. Manually analyzing and labeling 3357 to 5534 whale images could demand around 1328 to 2016 hours \cite{boulent2023scaling}. Gathering aerial images through field operations adds to expenses.  Additionally, labeling objects of interest in dynamic and complex maritime environments poses considerable difficulties~\cite{karapetyan2021human}. Hence, there is a pressing need for alternative methods to generate large-scale datasets rapidly, encompassing a wide variety of objects.

To overcome the scarcity of datasets in maritime environments, we introduce a simulation platform called \textbf{\textit{SeaDroneSim2}} to generate synthetic data and enhance object detection quality. By leveraging our simulation suit \textit{SeaDroneSim2}, synthetic aerial and satellite images can be created to replicate a wide range of objects and environmental conditions. Customizable virtual scenes can be rendered to replicate genuine maritime situations, expediting the creation of varied datasets. Synthetic data generation allows for variations in lighting conditions, altitudes, viewing angles, watercolors, and more. This provides a large variety of training examples that strengthen object detection and tracking systems. \textit{SeaDroneSim2} overcomes the challenges of manual data collection and facilitates the creation of large-scale datasets encompassing a variety of objects, including whales.

Our main contributions are as follows:
\begin{itemize}
\item We have made improvements to our novel simulation suit for generating aerial and satellite images for the maritime environment, adding enhanced functionality for noise and water properties. 
\item We conducted experiments to evaluate the synthetic datasets and compare performances.
\item We open-source \textit{SeaDronesSim2} and dataset associated with this work to accelerate further research. To the best of our knowledge, we are among the first to share the segmentation labeling for whale detection. 
\item We proposed a complete pipeline for autonomously generating aerial and satellite maritime images for objects of interest and detecting them. 
\end{itemize}
The rest of this paper is organized as follows: We first place this work in the context of related works in Sec.~\ref{section:related_work}. Then, we describe the proposed simulation which is used to create photo-realistic images in Sec.~\ref{section:problem_formulation}. We then present some quantitative and qualitative evaluations of our approach in Sec.~\ref{section:Experiments_and_results}. We conclude our work in Sec.~\ref{sec:conclusion} with parting thoughts on future work.
% In the following section, we will contextualize our work and place it within the framework of related studies.

\section{RELATED WORK}
\label{section:related_work}

This section reviews datasets, object detection for maritime environments, whale detection studies, and simulations in the maritime and aerial domains.

\subsection{Datasets and Object Detection for Maritime Environments}
The development of advanced computer vision algorithms necessitates access to extensive datasets, particularly in the context of maritime environments. While existing datasets predominantly center around synthetic aperture radar satellite imagery for remote sensing tasks, a growing trend is directing attention toward Very High-Resolution (VHR) images for object detection within these environments \cite{kapoor2023deep,gonccalves2020sealnet}. Noteworthy contributions include Gallego's \cite{gallego2018automatic} autonomous ship detection method using aerial images, and Li's \cite{li2018hsf} dataset featuring Google Earth and UAV-based images for ship detection. Lygouras et al \cite{lygouras2019unsupervised} focused on human detection with UAV-based images, albeit with dataset limitations. Kiefer et al \cite{kiefer2022leveraging} explored maritime and terrestrial images for boat and people detection. UAV-based dataset from Varga et al \cite{varga2022seadronessee} for water object recognition is noteworthy, though its applicability for SAR tasks might be constrained due to its limited object class coverage. 

In the realm of whale monitoring, Very High-Resolution (VHR) images have garnered substantial attention \cite{guirado2019whale,green2023gray,borowicz2019aerial} as a viable alternative to conventional techniques such as ship-based or acoustic-based monitoring \cite{torterotot2023long,hodul2023individual,aulich2019fin}. This shift highlights the growing preference for VHR images and their potential to offer insights into whale populations, behaviors, and associated risks. Deep neural networks have emerged as a pivotal tool in multiple endeavors to detect whales\cite{kapoor2023deep,green2023gray,bogucki2019applying}, harnessing their prowess in analyzing both visual and auditory data.
Boulent et al \cite{boulent2023scaling} proposed a human-in-the-loop approach that combines automation and biologist expertise, creating an AI-assisted annotation tool for whale monitoring. This demonstrates the potential of deep learning to enhance efficiency and accuracy in analyzing whale images, with implications for management and conservation policies.
% \begin{figure*}[ht!]
% \includegraphics[width=\textwidth]{./figures/huge_image_of_combinations.jpg}
% \centering
% \vspace{-3mm}
% \caption{These are the synthetic images generated from \textbf{\textit{SeaDroneSim2}}. In the first row, the first four images showcase the increasing turbidity of the water, and the last four images depict varying lighting conditions. In the second row, the first four images display different watercolors, while the last four images exhibit increasing noise from the satellite images. In the third row, the first four images demonstrate varying altitudes, while the last four images illustrate different whale positions, including lodging, spyhopping, and submerging.}
% \label{fig:different_features}
% \vspace{-3mm}
% \end{figure*}
\subsection{Simulation}
In our approach to detection tasks, we draw inspiration from the concept of detecting Remotely Operated Vehicle(ROV) \cite{lin2023seadronesim}, oysters \cite{lin2022oystersim,lin2023oysternet} and propellers\cite{sanket2021prgflow}, which utilize 3D models of the objects to generate synthetic data. Similarly, we utilize a 3D model of the whale to create a maritime dataset specifically for whale detection, addressing the lack of large-scale datasets in this domain.

When large-scale datasets are lacking for robotics tasks, research groups have developed simulations to meet their needs. One of the most relevant works,  RarePlanes\cite{shermeyer2021rareplanes}, focuses on utilizing synthetic images to detect airplanes in very high-resolution (VHR) images. Simulators in the aerial and maritime domains often focus on drone control for safety operations\cite{cox2007use} and rapid control\cite{velasco2020open}. Some examples(Abujob \cite{abujoub2018unmanned}) include simulations for verifying algorithms related to landing drones on ships with motion prediction. While similar simulators like the Matlab UAV Toolbox\cite{matlabuav} exist, they primarily focus on terrestrial applications and lack ground truth segmentation for the objects of interest.

With the ultimate goal of developing an autonomous aerial and satellite surveillance system, we recognize the importance of object detection methods. Due to the scarcity of large-scale datasets for the aerial and satellite of the maritime environment and limited literature on this topic, we are pursuing an alternative approach by generating datasets through synthetic image generation. To the best of our knowledge, we are among the first to propose a simulator called \textit{SeaDroneSim2} for generating aerial and satellite datasets of the maritime environment and using them for object detection. Details of the \textit{SeaDroneSim2} will be described in the following section.
\section{System Description}

\label{section:problem_formulation}
\textbf{\textit{SeaDroneSim2}} is built based on the Blender$^{\text{TM}}$ \cite{blender} game engine. After creating the simulated environment, the object of interest is incorporated to generate a synthetic dataset for training a Neural Network in object detection. As depicted in Fig.~\ref{fig:overview}), the chosen object and various parameters for the maritime environment(like water texture, lighting, and random objects) are inputted into \textit{SeaDroneSim2}. The open-source tool then produces a training dataset along with corresponding ground truth masks. This generated data facilitates and enhances the development of a detection network for recognizing the  object of interest. In the remainder of this section, we will go through some of the details of this image generation and object detection pipeline. 

For the maritime object detection application of \textit{SeaDroneSim2}, the Neural Network must be trained to detect a range of shapes and colors for specific objects. It should also account for diverse oceanic variables, such as water texture, watercolor (including different levels of turbidity),and lighting conditions. Training the detection network to be effective for these varying environments requires large training datasets, which, as aforementioned, are often costly in nature~\cite{cost}. At the time of writing, there are very limited training datasets for maritime object detection, one of which is a dataset from the British Antarctic Survey\cite{cubaynes2022whales}. Existing datasets, even the one from the British Antarctic Survey, lack ground truth masks for image segmentation of the targeted object. Therefore, images must be labeled by hand to train the Neural Network. Here, \textit{SeaDroneSim2} proves advantageous as it generates accurate ground truth image masks for objects within the simulation.
\subsection{Object of Interest}
Generating realistic synthetic environments relies heavily on an accurate 3D model of the object of interest. The selection of the object of interest is based on two key factors: the increasing conservation efforts for marine life, including whales, and the high costs associated with obtaining state-of-the-art datasets. In addition, the use of whales as our primary object of interest is also easily translatable to other marine creatures, such as dolphins and sea lions, due to their presence near the surface of the ocean.

As depicted in Fig.~\ref{fig:illustrative_example}, the 3D whale model closely resembles an actual whale when observed from a satellite or aerial perspective. Although some differences in detail are evident, the 3D model offers a realistic portrayal of whales in satellite imagery.

\begin{figure}[t!]
		\centering      
		\includegraphics[width=0.95\linewidth]{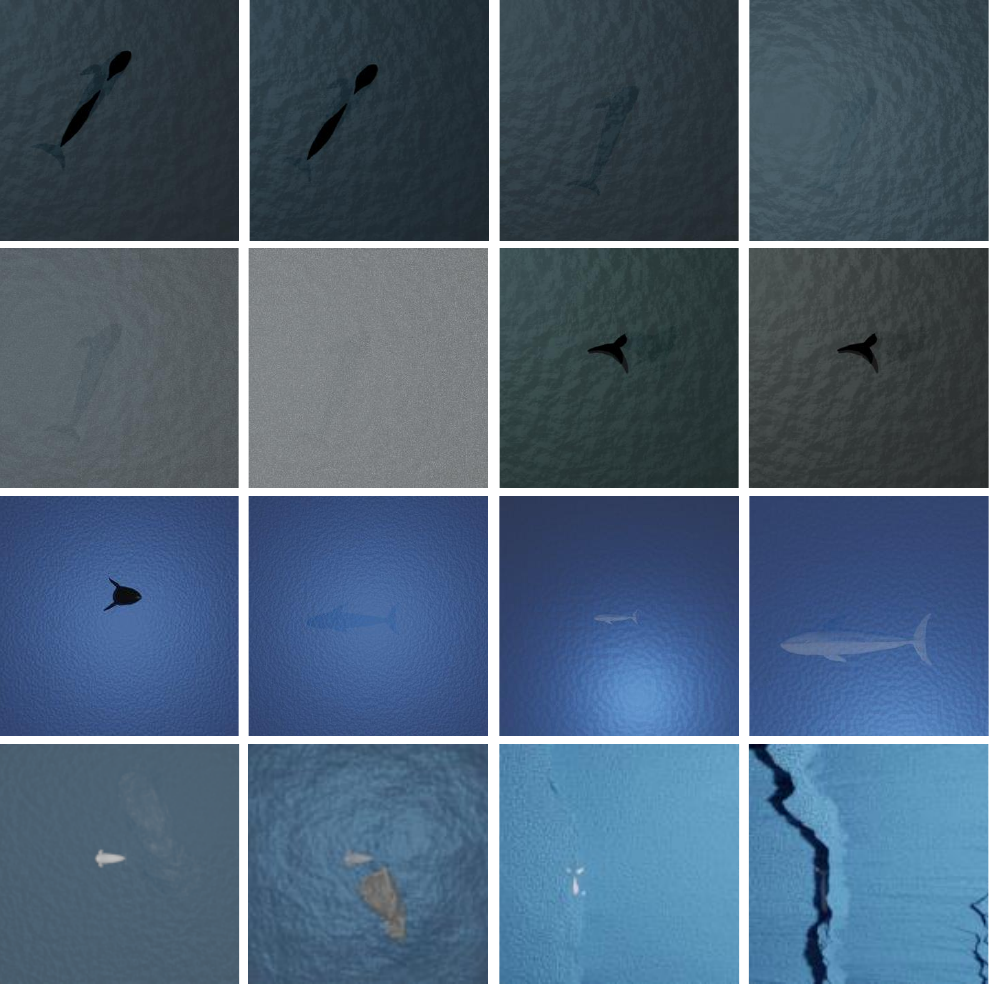}
		\caption{These are the synthetic images generated from \textbf{\textit{SeaDroneSim2}}. In the first row, the first two images showcase the increasing turbidity of the water, and the last two images depict varying lighting conditions. In the second row, the first two images display different watercolors, while the last two images exhibit increasing noise from the satellite images. In the third row, the first two images demonstrate varying altitudes, while the last two images illustrate different whale positions, including lodging, spyhopping, and submerging. In the last row, the first two images demonstrate synthetic images with different water waves. while the last two images illustrate different rocks and hills.}
		\label{fig:different_features}
  \vspace{-4mm}
	\end{figure}

Our synthetic data generation tool offers parameterization options specific to the object of interest. This includes the capability to rotate the object along three axes and move the object across three dimensions, as demonstrated in the first two images in the third row in Fig.~\ref{fig:different_features}. In addition to the physical location of the object of interest, customizable textures can also be applied to the object to simulate different environments, including different species of marine animals and marine objects(in the third row in Fig.~\ref{fig:different_features}). 
For each synthetically generated image of the object of interest,  \textit{SeaDroneSim2} provides functions to generate image masks. The masks aid in image segmentation and alleviate the requirement for manual image labeling, which can be labor-intensive.
Moreover, \textit{SeaDroneSim2} can generate 1000 training images (with resolution 140x140), with their masks, in about 25 minutes. This value varies depending on the resolution of the images being generated.

% \begin{figure}[t!]
% 		\centering      
% 		\includegraphics[width=\linewidth]{figures/watercolor_and_turbidity.pdf}
% 		\caption{Different water colors and turbidity levels.}
% 		\label{fig:water_effect}
%   \vspace{-4mm}
% 	\end{figure}
\subsection{Water Volume}
\textit{SeaDroneSim2} uses Blender$^{\text{TM}}$'s render engine CYCLES to create the synthetic oceanic environment. The render engine uses path-tracing of the light object to generate image renders. As a result, through several customization mechanisms, we can attain realistic oceanic environments for different watercolors, textures, and turbidity levels. 

A notable advantage of \textit{SeaDroneSim2} is its automated data generation process, requiring minimal user interaction. By utilizing a local installation of Blender$^{\text{TM}}$, the tool generates a Blender$^{\text{TM}}$ file containing an oceanic plane equipped with pre-configured lighting and camera placements.

\textit{SeaDroneSim2} provides utilities for both customized and pre-built implementations, enabling users to fine-tune maritime environments. For instance, users can change water color using specific \textit{RGB} value ranges(in the last two images in the second-row Fig.~\ref{fig:different_features}), generating corresponding images and masks. The default implementation offers 1030 different colors, ranging from blue to green water shades. Users can also adjust water turbidity, affecting object visibility. The first two images in the first-row Fig.~\ref{fig:different_features} show increasing turbidity which decreases the whale's visibility.

In addition, water texture is another key parameter in the tool suite. As we can see in Fig.~\ref{fig:different_features}, there are several parameterization options for water texture, depending on different factors like detail, dimension, scale, metallic, \textit{lacunarity}, and strength. First, the water's detail fine-tunes the roughness of the water, while scale and strength play the most important factors in defining the roughness, with a larger scale resulting in calmer waters. The water's dimension and lacunarity compress or expand the water's patterns to also affect the roughness. Finally, the water's metallic reflects a varying level of refraction of the water itself.

Next, considering the prevalence of white noise in many available satellite oceanic datasets, there is often an element of white noise in datasets. \textit{SeaDroneSim2} also can simulate and render the noise through Blender$^{\text{TM}}$. Employing both \textit{White Noise} and \textit{Gaussian Noise}, the tool replicates different noise levels within images, as demonstrated in the first two images of the second row in Fig.~\ref{fig:different_features}. Similarly,  \textit{SeaDroneSim2} also enables varying levels of lighting of the ocean, mimicking varying levels of sunlight, depicted in the last two images in the first row of Fig.~\ref{fig:different_features}

Lastly, we incorporate the capability to simulate diverse wave characteristics in oceanic settings, as illustrated in the last two images of the last row in Fig.~\ref{fig:different_features}. These simulated waves encompass variations in height, tilt, sharpness, and textures, offering a range of parameterized functions.

\subsection{Objects of Non-Interest}
Moreover, for a comprehensive and immersive environment, enhancements can be made to the surroundings beyond the water. This could involve adding buoys and markers, seabed features, rocks, reefs, or other relevant elements to provide context and realism to the oceanic setting.  Such inclusions can contribute to a more realistic simulation and better represent the intricacies of maritime environments.
As depicted in the first two images in the last row of Fig.~\ref{fig:different_features}, an additional rock has been included on the seafloor, showcasing how submerged objects can be incorporated into the simulation. The differing visibility levels and rock height demonstrate how water turbidity influences the clarity of submerged objects.
\subsection{Aerial Image}
One of our goals is to simulate aerial images in \textit{SeaDroneSim2}. While other camera angles are easily achievable through Blender$^{\text{TM}}$'s capabilities, most images take advantage of a simple aerial view. However, there does exist variance in camera altitude, which yields larger objects of interest at lower altitudes and smaller objects of interest at higher altitudes. 
The last two images in the third row Fig.~\ref{fig:different_features} illustrate this variance, showcasing how the camera can be parameterized based on the object's specifications, while typically being fixated on the object of interest by default.
While the majority of training images maintain a resolution of 140 $\times$ 140 pixels, the image resolution can extend to 30,000 $\times$ 30,000 pixels, depending on the computational capabilities of the underlying hardware.

In summary, by combining realistic rendering of water properties, such as transparency and wave dynamics, with the integration of diverse underwater elements, \textit{SeaDroneSim2}can serve as a valuable tool for exploring maritime situations, enhancing and evaluating detection algorithms, contributing to cetacean conservation, and various other applications.
\begin{figure*}[ht!]
\includegraphics[width=\textwidth]{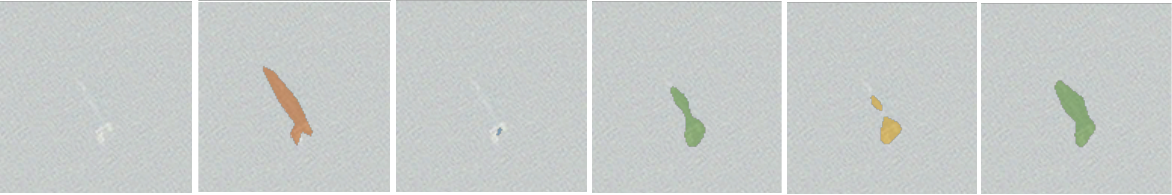}
\centering
\vspace{-3mm}
\caption{From left to right: Sample real input image, ground truth, segmentation result using Unet without synthetic augmented
real data, segmentation result using Unet with synthetic augmented
real data, segmentation result using FPN without synthetic augmented
real data, segmentation result using FPN with synthetic augmented
real data. All networks here are trained with only 10\% of real data. }
\label{fig:comparison}
\vspace{-3mm}
\end{figure*}

\section{Experiments And Results}
\label{section:Experiments_and_results}
First, we describe the dataset we used in these experiments. Then we compared the results obtained by two different segmentation networks of detecting the whales utilizing datasets generated from \textit{SeaDroneSim2}. 
% \subsection{Whale Detection}
% We first describe the datasets used in this subsection. Then, we compare the results obtained by two different segmentation networks with various dataset sizes. 
\subsection{Synthetic Dataset}

During our rendering process, we produced a total of 2000 synthetic images to construct a diverse and representative dataset. To mimic real-world conditions, we introduced variations in lighting, altitudes, orientation, water color, Gaussian noise levels, water turbidity, and wave patterns. These modifications were implemented to approximate the realistic characteristics commonly observed in satellite images, thereby creating a more photo-realistic environment for the network to learn from.

Moreover, for training the network to segment whales in diverse body orientations of the whale, we incorporated synthetic whale instances with varying body orientations of the whale in the simulations. Using this approach, we enforced the network to segment whales more accurately, regardless of their body orientation or appearance.

\subsection{Real Dataset}
To obtain whale images from the satellite view, we accessed a collection of 633 images through the UK Polar Data Centre (PDC) \cite{cubaynes2022whales}. These images played a crucial role in evaluating the effectiveness of our approach. However, during our dataset review, we noticed that the majority of these images only captured a small portion of the whale above the sea surface, and in some cases, only a splash of water from the whales.

Due to the unique characteristics of these satellite images, we conducted a meticulous examination to identify suitable ones that could be included in our test dataset. After careful selection, we were able to compile a set of 508 images that met our criteria and could be effectively utilized for testing and evaluating our approach.

Although this subset of real satellite images may have limitations in terms of whale visibility and coverage, it provides us with valuable test data to assess the performance and robustness of our method in segmenting and identifying whales under challenging scenarios.
\subsubsection{Evaluation Metrics}
To evaluate with the real dataset, we use the Intersection over Union (IoU) which is a common evaluation metric used to assess the performance of image segmentation algorithms. It measures the similarity between the predicted segmentation and the ground truth segmentation.
The term \textit{Intersection} refers to the region where the predicted segmentation and the ground truth segmentation overlap. On the other hand, \textit{Union} encompasses the entire area covered by both segments, whether overlapping or not. To assess the model's performance, we define the success and Detection Rate (DR) for each cluster based on the IoU as
\begin{equation}
    \label{matric}
        \text{Success}:=IoU\ge \tau; \text{DR}=\frac{\text{TP}}{\text{TP+FN}}
\end{equation}

where TP represents the true positive count, which corresponds to the number of correctly segmented instances by the model, and FN represents the false negative count, signifying the number of instances that were present in the ground truth but were not segmented by the model. We assign DR with two different thresholds $\tau$ of 0.5 and 0.6, which are denoted as $DR_{50}$mn  and $DR_{60}$ in Table~\ref{tab:segmentation_result}. Finally, we conduct a comprehensive analysis by computing a series of results for varying percentages of real data size, allowing us to make a thorough comparison between the method with and without the incorporation of \textit{SeaDroneSim2}.

\subsection{Experimental Results}

During our testing phase, we evaluated two types of networks: Unet \cite{ronneberger2015u} and Feature Pyramid Networks (FPN) \cite{lin2017feature}. For both networks, we used a learning rate of 0.001 with decay. The optimization process utilized the Adam optimizer in conjunction with the Jaccard loss \cite{bertels2019optimizing} as our loss function.

Before training, we augmented the dataset using various techniques, including a rotation range of 90 degrees, width shift range of 0.3, height shift range of 0.3, shear range of 0.5, zoom range of 0.3, horizontal flip, and vertical flip.
These data augmentation techniques are applied to the images to artificially increase the dataset's diversity and expose the model to various transformations that may occur in real-world scenarios. By incorporating these augmentations, we aim to enhance the model's ability to generalize and improve its performance on unseen data during training.

Throughout the training process, we used a batch size of 32 and conducted training for 100 epochs. To ensure reliable and consistent results, each training was performed at least 10 times (tests), and the best result was selected for further analysis and comparison.
\begin{table}[t!]
\caption{Comparison of Semantic Segmentation Results with the Two Different Networks}
\centering
% |p{1.2cm}|p{1.1cm}|p{1.7cm}|p{1.1cm}|p{1.6cm}|
\begin{tabular}{llll}
\toprule
% \multicolumn{5}{c}{Testing accuracy} \\
% \hline
Method & $DR_{50}$ & $DR_{60}$ & Real Data size\\
\hline
\textit{Unet}  & 0.615  & 0.488 & 10\% \\
\textit{Unet +SeaDroneSim2} &  \textbf{0.710} & \textbf{0.512} & 10\%\\
\hline
\textit{Unet}  &  0.849  & 0.687 & 50\% \\
\textit{Unet +SeaDroneSim2} &  0.833 & \textbf{0.710} & 50\%\\

\hline
\textit{FPN}  & 0.679  &  0.476 & 10\% \\
\textit{FPN +SeaDroneSim2} & \textbf{0.746} & \textbf{0.551} & 10\%\\

\hline
\textit{FPN}  & 0.861  &  0.714 & 50\% \\
\textit{FPN +SeaDroneSim2} & 0.861 & 0.683 & 50\%\\

\bottomrule
\end{tabular}
\label{tab:segmentation_result}
\end{table}
By conducting these extensive tests on both Unet and FPN networks, we aimed to determine the usability of our rendered synthetic image and identify the most suitable model for \textit{SeaDroneSim2} application.

Given the significant time and financial expenses involved in collecting real datasets, our strategy focuses on minimizing the reliance on real data for training. To achieve this goal, we performed tests using different proportions of the real dataset, specifically 10\% and 50\% of the available real data. The remaining 50\% of the real dataset (our held real data set) is reserved for testing purposes.

Both the Unet and FPN methods were tested, and we obtained $DR_{50}$ and $DR_{60}$ scores using only 10\% of the real dataset, which serves as our baseline for the task. For the Unet method, the $DR_{50}$ and $DR_{60}$ scores were 0.615 and 0.488, respectively. As for the FPN method, the $DR_{50}$ and $DR_{60}$ scores were 0.679 and 0.476, respectively.  The results are tabulated in Table~\ref{tab:segmentation_result}
After training the model solely on synthetic datasets and conducting the testing, we observed that the results were lower than the baseline. We think both networks have learned to recognize whales in the synthetic domain but the sim-to-real domain transfer is lacking in this case. 

We achieve better results when including the synthetic dataset rendered by \textit{SeaDroneSim2}. The $DR_{50}$ and $DR_{60}$ results from `Unet+\textit{SeaDroneSim}' (Unet with synthetic augmented real data) are 0.71 and 0.512 against the human-labeled ground truth which is 15.4\% and 4.91\% better than just using a 10\% real dataset for training.
The $DR_{50}$ and $DR_{60}$ results from "FPN+\textit{SeaDroneSim}" (FPN with synthetic augmented real data) are 0.746 and 0.551 against the human-labeled ground truth which is 9.86\% and 15.7\% better than just using a 10\% real dataset for training. Moreover, the $DR_{60}$ results from `Unet + SeaDroneSim (Unet with synthetic augmented real data) are 0.71, which is 3.34\% better than just using a 50\% real dataset for training. 

As depicted in Fig.~\ref{fig:comparison}, the underwater submerged portion of the body of the whale would be overlooked if the network were not trained with synthetic augmented real data. Consequently, this situation results in numerous false negatives. Our synthetic data includes numerous samples where either a portion of the whale's body or the entire body is submerged in water. Thus, training the network using real data augmented with our synthetic data leads to a substantial reduction in false negatives. This improvement leads to more accurate predictions and a higher success rate even in challenging scenarios.

Augment the dataset with synthetic does not always guarantee improvement, as evidenced by the slight drop in the $DR_{60}$ result from both "\textit{FPN+SeaDroneSim}" and $DR_{50}$ result from"\textit{Unet+SeaDroneSim}" when using 50\% of the real dataset in Table.~\ref{tab:segmentation_result}. 
However, this drop could be attributed to the fact that our dataset is relatively small, which leads to limited robustness in handling diverse real-world scenarios for whale detection.
\section{Conclusions and Future Work}
\label{sec:conclusion}

In this work, we discussed how to utilize the capability of a render engine and built a simulation environment for whale detection. We discussed the implementation details of using Blender$^{\text{TM}}$ for generating synthetic datasets for object detection. We then compared our detection results with the usage of different synthetic datasets generated from the simulation. These results highlight that collecting real images is challenging for data-critical applications. It is possible to use 3D models of the object to create photorealistic images in a simulation environment that will successfully detect objects in a specific domain. This work is among the first to build a maritime object simulation focusing on object detection with particular emphasis on whale detection. 

In the future, we aim to enhance the capabilities and functionalities of \textit{SeaDroneSim2} and compare our results with additional datasets for different objects of interest, including coral reefs and oysters. In addition to our ongoing efforts, we are dedicated to creating specialized datasets that focus on maritime objects, with a particular emphasis on aiding cetacean scientists in detecting and monitoring whales under the ice. By providing these tailored datasets, we aim to equip researchers with the necessary tools to effectively study and protect whales in challenging icy environments. 
%%%%%%%%%%%%%%%%%%%%%%%%%%%%%%%%%%%%%%%%%%%%%%%%%%%%%%%%%%%%%%%%%%%%%%%%%%%%%%%%

\section*{ACKNOWLEDGMENT}

The authors would like to thank British Antarctic Survey (BAS) for publishing the real whale datasets online with open access to the public. This work is supported by "Transforming Shellfish Farming with Smart Technology and Management Practices for Sustainable Production" grant no. 2020-68012-31805/project accession no. 1023149 from the USDA National Institute of Food and Agriculture.

%%%%%%%%%%%%%%%%%%%%%%%%%%%%%%%%%%%%%%%%%%%%%%%%%%%%%%%%%%%%%%%%%%%%%%%%%%%%%%%%
\bibliographystyle{template/IEEEtran}
\bibliography{refs}

\begin{thebibliography}{10}
\providecommand{\url}[1]{#1}
\csname url@rmstyle\endcsname
\providecommand{\newblock}{\relax}
\providecommand{\bibinfo}[2]{#2}
\providecommand\BIBentrySTDinterwordspacing{\spaceskip=0pt\relax}
\providecommand\BIBentryALTinterwordstretchfactor{4}
\providecommand\BIBentryALTinterwordspacing{\spaceskip=\fontdimen2\font plus
\BIBentryALTinterwordstretchfactor\fontdimen3\font minus
  \fontdimen4\font\relax}
\providecommand\BIBforeignlanguage[2]{{%
\expandafter\ifx\csname l@#1\endcsname\relax
\typeout{** WARNING: IEEEtran.bst: No hyphenation pattern has been}%
\typeout{** loaded for the language `#1'. Using the pattern for}%
\typeout{** the default language instead.}%
\else
\language=\csname l@#1\endcsname
\fi
#2}}

\bibitem{chen2020imaging}
P.-C. Chen, Y.-C. Chiang, and P.-Y. Weng, ``Imaging using unmanned aerial
  vehicles for agriculture land use classification,'' \emph{Agriculture},
  vol.~10, no.~9, p. 416, 2020.

\bibitem{raj2020precision}
R.~Raj, S.~Kar, R.~Nandan, and A.~Jagarlapudi, ``Precision agriculture and
  unmanned aerial vehicles (uavs),'' in \emph{Unmanned Aerial Vehicle:
  Applications in Agriculture and Environment}.\hskip 1em plus 0.5em minus
  0.4em\relax Springer, 2020, pp. 7--23.

\bibitem{tsouros2019review}
D.~C. Tsouros, S.~Bibi, and P.~G. Sarigiannidis, ``A review on uav-based
  applications for precision agriculture,'' \emph{Information}, vol.~10,
  no.~11, p. 349, 2019.

\bibitem{adade2021unmanned}
R.~Adade, A.~M. Aibinu, B.~Ekumah, and J.~Asaana, ``Unmanned aerial vehicle
  (uav) applications in coastal zone management—a review,''
  \emph{Environmental Monitoring and Assessment}, vol. 193, no.~3, pp. 1--12,
  2021.

\bibitem{albanese2021sardo}
A.~Albanese, V.~Sciancalepore, and X.~Costa-P{\'e}rez, ``Sardo: An automated
  search-and-rescue drone-based solution for victims localization,'' \emph{IEEE
  Transactions on Mobile Computing}, vol.~21, no.~9, pp. 3312--3325, 2021.

\bibitem{cubaynes2019whales}
H.~C. Cubaynes, P.~T. Fretwell, C.~Bamford, L.~Gerrish, and J.~A. Jackson,
  ``Whales from space: four mysticete species described using new vhr satellite
  imagery,'' \emph{Marine Mammal Science}, vol.~35, no.~2, pp. 466--491, 2019.

\bibitem{gonccalves2020sealnet}
B.~C. Gon{\c{c}}alves, B.~Spitzbart, and H.~J. Lynch, ``Sealnet: A
  fully-automated pack-ice seal detection pipeline for sub-meter satellite
  imagery,'' \emph{Remote Sensing of Environment}, vol. 239, p. 111617, 2020.

\bibitem{karapetyan2021robot}
N.~Karapetyan, ``Robot area coverage path planning in aquatic environments,''
  Ph.D. dissertation, University of South Carolina, 2021.

\bibitem{cubaynes2022whales}
H.~C. Cubaynes and P.~T. Fretwell, ``Whales from space dataset, an annotated
  satellite image dataset of whales for training machine learning models,''
  \emph{Scientific Data}, vol.~9, no.~1, p. 245, 2022.

\bibitem{cost}
\BIBentryALTinterwordspacing
P.~Mumby, E.~Green, A.~Edwards, and C.~Clark, ``The cost-effectiveness of
  remote sensing for tropical coastal resources assessment and management,''
  \emph{Journal of Environmental Management}, vol.~55, no.~3, pp. 157--166,
  Mar. 1999. [Online]. Available: \url{https://doi.org/10.1006/jema.1998.0255}
\BIBentrySTDinterwordspacing

\bibitem{boulent2023scaling}
J.~Boulent, B.~Charry, M.~M. Kennedy, E.~Tissier, R.~Fan, M.~Marcoux, C.~A.
  Watt, and A.~Gagn{\'e}-Turcotte, ``Scaling whale monitoring using deep
  learning: A human-in-the-loop solution for analyzing aerial datasets,''
  \emph{Frontiers in Marine Science}, vol.~10, p. 1099479, 2023.

\bibitem{karapetyan2021human}
N.~Karapetyan, J.~V. Johnson, and I.~Rekleitis, ``Human diver-inspired visual
  navigation: Towards coverage path planning of shipwrecks,'' \emph{Marine
  Technology Society Journal}, vol.~55, no.~4, pp. 24--32, 2021.

\bibitem{kapoor2023deep}
S.~Kapoor, M.~Kumar, and M.~Kaushal, ``Deep learning based whale detection from
  satellite imagery,'' \emph{Sustainable Computing: Informatics and Systems},
  vol.~38, p. 100858, 2023.

\bibitem{gallego2018automatic}
A.-J. Gallego, A.~Pertusa, and P.~Gil, ``Automatic ship classification from
  optical aerial images with convolutional neural networks,'' \emph{Remote
  Sensing}, vol.~10, no.~4, p. 511, 2018.

\bibitem{li2018hsf}
Q.~Li, L.~Mou, Q.~Liu, Y.~Wang, and X.~X. Zhu, ``Hsf-net: Multiscale deep
  feature embedding for ship detection in optical remote sensing imagery,''
  \emph{IEEE Transactions on Geoscience and Remote Sensing}, vol.~56, no.~12,
  pp. 7147--7161, 2018.

\bibitem{lygouras2019unsupervised}
E.~Lygouras, N.~Santavas, A.~Taitzoglou, K.~Tarchanidis, A.~Mitropoulos, and
  A.~Gasteratos, ``Unsupervised human detection with an embedded vision system
  on a fully autonomous uav for search and rescue operations,'' \emph{Sensors},
  vol.~19, no.~16, p. 3542, 2019.

\bibitem{kiefer2022leveraging}
B.~Kiefer, D.~Ott, and A.~Zell, ``Leveraging synthetic data in object detection
  on unmanned aerial vehicles,'' in \emph{2022 26th International Conference on
  Pattern Recognition (ICPR)}.\hskip 1em plus 0.5em minus 0.4em\relax IEEE,
  2022, pp. 3564--3571.

\bibitem{varga2022seadronessee}
L.~A. Varga, B.~Kiefer, M.~Messmer, and A.~Zell, ``Seadronessee: A maritime
  benchmark for detecting humans in open water,'' in \emph{Proceedings of the
  IEEE/CVF Winter Conference on Applications of Computer Vision}, 2022, pp.
  2260--2270.

\bibitem{guirado2019whale}
E.~Guirado, S.~Tabik, M.~L. Rivas, D.~Alcaraz-Segura, and F.~Herrera, ``Whale
  counting in satellite and aerial images with deep learning,''
  \emph{Scientific reports}, vol.~9, no.~1, p. 14259, 2019.

\bibitem{green2023gray}
K.~M. Green, M.~K. Virdee, H.~C. Cubaynes, A.~I. Aviles-Rivero, P.~T. Fretwell,
  P.~C. Gray, D.~W. Johnston, C.-B. Sch{\"o}nlieb, L.~G. Torres, and J.~A.
  Jackson, ``Gray whale detection in satellite imagery using deep learning,''
  \emph{Remote Sensing in Ecology and Conservation}, 2023.

\bibitem{borowicz2019aerial}
A.~Borowicz, H.~Le, G.~Humphries, G.~Nehls, C.~H{\"o}schle, V.~Kosarev, and
  H.~J. Lynch, ``Aerial-trained deep learning networks for surveying cetaceans
  from satellite imagery,'' \emph{PloS one}, vol.~14, no.~10, p. e0212532,
  2019.

\bibitem{torterotot2023long}
M.~Torterotot, F.~Samaran, and J.-Y. Royer, ``Long-term acoustic monitoring of
  nonstereotyped blue whale calls in the southern indian ocean,'' \emph{Marine
  Mammal Science}, vol.~39, no.~2, pp. 594--610, 2023.

\bibitem{hodul2023individual}
M.~Hodul, A.~Knudby, B.~McKenna, A.~James, C.~Mayo, M.~Brown, D.~Durette-Morin,
  and S.~Bird, ``Individual north atlantic right whales identified from
  space,'' \emph{Marine Mammal Science}, vol.~39, no.~1, pp. 220--231, 2023.

\bibitem{aulich2019fin}
M.~G. Aulich, R.~D. McCauley, B.~J. Saunders, and M.~J. Parsons, ``Fin whale
  (balaenoptera physalus) migration in australian waters using passive acoustic
  monitoring,'' \emph{Scientific reports}, vol.~9, no.~1, p. 8840, 2019.

\bibitem{bogucki2019applying}
R.~Bogucki, M.~Cygan, C.~B. Khan, M.~Klimek, J.~K. Milczek, and M.~Mucha,
  ``Applying deep learning to right whale photo identification,''
  \emph{Conservation Biology}, vol.~33, no.~3, pp. 676--684, 2019.

\bibitem{lin2023seadronesim}
X.~Lin, C.~Liu, A.~Pattillo, M.~Yu, and Y.~Aloimonous, ``Seadronesim:
  Simulation of aerial images for detection of objects above water,'' in
  \emph{Proceedings of the IEEE/CVF Winter Conference on Applications of
  Computer Vision}, 2023, pp. 216--223.

\bibitem{lin2022oystersim}
X.~Lin, N.~Jha, M.~Joshi, N.~Karapetyan, Y.~Aloimonos, and M.~Yu, ``Oystersim:
  Underwater simulation for enhancing oyster reef monitoring,'' in \emph{OCEANS
  2022, Hampton Roads}.\hskip 1em plus 0.5em minus 0.4em\relax IEEE, 2022, pp.
  1--6.

\bibitem{lin2023oysternet}
X.~Lin, N.~J. Sanket, N.~Karapetyan, and Y.~Aloimonos, ``Oysternet: Enhanced
  oyster detection using simulation,'' in \emph{2023 IEEE International
  Conference on Robotics and Automation (ICRA)}.\hskip 1em plus 0.5em minus
  0.4em\relax IEEE, 2023, pp. 5170--5176.

\bibitem{sanket2021prgflow}
N.~J. Sanket, C.~D. Singh, C.~Ferm{\"u}ller, and Y.~Aloimonos, ``Prgflow:
  Unified swap-aware deep global optical flow for aerial robot navigation,''
  \emph{Electronics Letters}, vol.~57, no.~16, pp. 614--617, 2021.

\bibitem{shermeyer2021rareplanes}
J.~Shermeyer, T.~Hossler, A.~Van~Etten, D.~Hogan, R.~Lewis, and D.~Kim,
  ``Rareplanes: Synthetic data takes flight,'' in \emph{Proceedings of the
  IEEE/CVF Winter Conference on Applications of Computer Vision}, 2021, pp.
  207--217.

\bibitem{cox2007use}
I.~Cox, J.~Howitt, and J.~Duncan, ``The use of simulation to de-risk maritime
  uav operations,'' SYSTEM ENGINEERING AND ASSESSMENT LTD BRISTOL (UNITED
  KINGDOM), Tech. Rep., 2007.

\bibitem{velasco2020open}
O.~Velasco, J.~Valente, P.~J. Alhama~Blanco, and M.~Abderrahim, ``An open
  simulation strategy for rapid control design in aerial and maritime drone
  teams: A comprehensive tutorial,'' \emph{Drones}, vol.~4, no.~3, p.~37, 2020.

\bibitem{abujoub2018unmanned}
S.~Abujoub, J.~Mcphee, C.~Westin, and R.~A. Irani, ``Unmanned aerial vehicle
  landing on maritime vessels using signal prediction of the ship motion,'' in
  \emph{OCEANS 2018 MTS/IEEE Charleston}.\hskip 1em plus 0.5em minus
  0.4em\relax IEEE, 2018, pp. 1--9.

\bibitem{matlabuav}
\BIBentryALTinterwordspacing
I.~The~MathWorks, \emph{UAV Toolbox}, Natick, Massachusetts, United State,
  2019. [Online]. Available: \url{https://www.mathworks.com/products/uav.html#}
\BIBentrySTDinterwordspacing

\bibitem{blender}
\BIBentryALTinterwordspacing
B.~O. Community, \emph{Blender - a 3D modelling and rendering package}, Blender
  Foundation, Stichting Blender Foundation, Amsterdam, 2018. [Online].
  Available: \url{http://www.blender.org}
\BIBentrySTDinterwordspacing

\bibitem{ronneberger2015u}
O.~Ronneberger, P.~Fischer, and T.~Brox, ``U-net: Convolutional networks for
  biomedical image segmentation,'' in \emph{International Conference on Medical
  image computing and computer-assisted intervention}.\hskip 1em plus 0.5em
  minus 0.4em\relax Springer, 2015, pp. 234--241.

\bibitem{lin2017feature}
T.-Y. Lin, P.~Doll{\'a}r, R.~Girshick, K.~He, B.~Hariharan, and S.~Belongie,
  ``Feature pyramid networks for object detection,'' in \emph{Proceedings of
  the IEEE conference on computer vision and pattern recognition}, 2017, pp.
  2117--2125.

\bibitem{bertels2019optimizing}
J.~Bertels, T.~Eelbode, M.~Berman, D.~Vandermeulen, F.~Maes, R.~Bisschops, and
  M.~B. Blaschko, ``Optimizing the dice score and jaccard index for medical
  image segmentation: Theory and practice,'' in \emph{Medical Image Computing
  and Computer Assisted Intervention--MICCAI 2019: 22nd International
  Conference, Shenzhen, China, October 13--17, 2019, Proceedings, Part II
  22}.\hskip 1em plus 0.5em minus 0.4em\relax Springer, 2019, pp. 92--100.

\end{thebibliography}

\end{document}